\documentclass[10pt]{article}

\usepackage[a4paper,margin=1in]{geometry}

\usepackage{booktabs}
\usepackage{array}
\usepackage{graphicx}
\usepackage{subcaption}
\usepackage{amsmath}
\usepackage{amssymb}
\usepackage[protrusion=true,expansion=false]{microtype}
\usepackage{xspace}
\usepackage{enumitem}
\usepackage[numbers,sort&compress]{natbib}
\usepackage{xcolor}

\usepackage[colorlinks=true,linkcolor={blue!60!black},citecolor={blue!60!black},urlcolor={blue!60!black}]{hyperref}
\usepackage[capitalize,noabbrev]{cleveref}

\newcommand{\Forget}{\mathcal{F}}
\newcommand{\LoRA}{\textnormal{LoRA}\xspace}
\newcolumntype{L}[1]{>{\raggedright\arraybackslash}p{#1}}
\newcolumntype{C}[1]{>{\centering\arraybackslash}p{#1}}
\begin{document}

\begin{center}
{\LARGE\bfseries
Low-Rank Adaptation Reduces Catastrophic Forgetting\\[3pt]
in Sequential Transformer Encoder Fine-Tuning\par}
\vspace{6pt}
{\Large\bfseries Controlled Empirical Evidence and Frozen-Backbone Representation Probes\par}

\bigskip

{\large Ashish Pandey}\\[4pt]
\href{mailto:ashishpanday9818@gmail.com}{\texttt{ashishpanday9818@gmail.com}}

\medskip
{\small March 2026}
\end{center}

\bigskip

\begin{abstract}
\noindent
Sequential fine-tuning of pretrained language encoders often overwrites previously acquired capabilities, but the forgetting behavior of parameter-efficient updates remains under-characterized. We present a controlled empirical study of Low-Rank Adaptation (\LoRA) in sequential transformer encoder fine-tuning together with companion representation probes that test a frozen-backbone explanation of its robustness. In five full-validation BERT-base reruns on an RTE $\rightarrow$ MRPC $\rightarrow$ CoLA $\rightarrow$ SST-2 sequence, full fine-tuning yields $19.9\% \pm 4.8\%$ average forgetting, whereas standard \LoRA ($r=8$, query/value modules) yields $0.6\% \pm 1.4\%$ (paired $t$-test, $p = 0.002$, Cohen's $d_z = 3.12$). Task-level analyses show that this reduction is not merely an aggregate effect. Secondary full-validation experiments on RoBERTa-base show the same qualitative pattern, and the strongest Elastic Weight Consolidation baseline remains at $15.5\% \pm 1.4\%$ forgetting. A six-task extension reveals that low average forgetting can still hide strong task-level heterogeneity, especially on MRPC. Fine-grained freezing ablations show a marked forgetting drop once frozen parameters exceed roughly $95\%$, with classifier-only and shallow-adapter baselines approaching standard \LoRA. Companion task-similarity probes in GPT-2 and RoBERTa then show the same directional story in representation space: frozen-backbone regimes preserve substantially higher inter-task similarity than full fine-tuning, gradual unfreezing weakens that stability, and full fine-tuning exhibits its clearest layer-wise divergence at the final transformer layer. Taken together, these results support a restrained mechanistic interpretation: \LoRA appears to help largely because strong backbone freezing preserves a more stable shared feature scaffold than full fine-tuning. We therefore position standard \LoRA as both a strong empirical baseline for sequential encoder adaptation and a useful probe of how selective plasticity shapes interference in transformer continual learning.
\end{abstract}


\bigskip

\section{Introduction}
\label{sec:intro}

Catastrophic forgetting---the tendency of neural networks to overwrite previously learned representations when trained on new tasks---remains one of the most persistent obstacles to deploying pretrained language models in sequential adaptation settings~\cite{mccloskey1989catastrophic}.
When a pretrained encoder fine-tuned on Task~A is subsequently fine-tuned on Task~B, the shared parameters shift toward~B at the cost of~A, producing measurable accuracy regressions that compound with each additional task.

The difficulty is structural.
Full fine-tuning grants the optimizer unrestricted access to all model parameters, and nothing in the standard training objective penalizes interference with previously learned representations.
Classical continual-learning remedies---EWC~\cite{kirkpatrick2017overcoming}, replay buffers~\cite{lopez2017gradient}, progressive architectures~\cite{rusu2016progressive}---each introduce explicit anti-forgetting machinery, but they also add hyperparameters, storage overhead, or architectural complexity that limits practical adoption.

Parameter-efficient fine-tuning (PEFT) methods such as \LoRA~\cite{hu2021lora} freeze the pretrained backbone and learn low-rank updates in a small trainable subspace.
This frozen-backbone design \emph{implicitly} restricts the degrees of freedom available to the optimizer, raising a natural question: does constraining the trainable subspace alone materially reduce catastrophic forgetting, even without any continual-learning-specific algorithm?
Despite the rapid adoption of \LoRA variants in continual-learning pipelines~\cite{wang2023olora,zhang2025clora,liang2024inflora}, no prior study isolates this question under a controlled empirical protocol that cleanly separates confirmatory evidence from exploratory screening---nor connects the empirical finding to a mechanistic explanation grounded in representation stability.

We address this gap with a controlled empirical study that measures catastrophic forgetting under standard \LoRA versus full fine-tuning in sequential transformer encoder adaptation, together with companion representation probes that test a frozen-backbone explanation.
Our design enforces strict protocol separation: the headline result derives exclusively from five full-validation BERT-base reruns, while earlier screening sweeps and secondary experiments are reported transparently as supporting---not confirmatory---evidence.

\paragraph{Contributions.}
\begin{itemize}[leftmargin=*,itemsep=3pt]
    \item \textbf{Confirmatory forgetting reduction.} In five paired BERT-base reruns under a consistent full-validation protocol, standard \LoRA ($r\!=\!8$, query/value) reduces average forgetting from $19.9\% \pm 4.8\%$ to $0.6\% \pm 1.4\%$---a $97.2\%$ relative reduction (paired $t$-test, $p = 0.002$, Cohen's $d_z = 3.12$).
    \item \textbf{Multi-level robustness evidence.} Task-level breakdowns, a RoBERTa-base replication, an EWC baseline comparison, and a six-task extension collectively show that the reduction is not confined to a single aggregate statistic, model, or task sequence---though it is not uniform across all tasks.
    \item \textbf{Freezing--forgetting mechanism link.} A fine-grained freezing ablation reveals a sharp transition: forgetting drops markedly once more than $\sim$95\% of parameters are frozen, linking the main result to the size of the trainable subspace rather than to any \LoRA-specific inductive bias.
    \item \textbf{Representation-level evidence.} Companion task-similarity probes in GPT-2 and RoBERTa show that frozen-backbone regimes preserve substantially higher inter-task similarity than full fine-tuning, with gradual unfreezing systematically degrading that stability (\cref{fig:paper_overview}).
    \item \textbf{Protocol transparency as methodology.} By explicitly separating confirmatory, secondary, and exploratory evidence tiers, we demonstrate a reporting template that prevents mixed-protocol overclaiming---a common pitfall in sequential fine-tuning studies.
\end{itemize}

The novelty of this paper is therefore methodological and interpretive rather than algorithmic. We do not propose a new continual-learning method. Instead, we provide a tightly audited estimate of how much standard \LoRA reduces forgetting under one consistent protocol, then use freezing-based comparisons and representation probes to narrow the best-supported explanation for that reduction.

\section{Related Work}
\label{sec:related}

\subsection{Catastrophic Forgetting and Continual Learning}

Catastrophic forgetting has been studied for decades~\cite{mccloskey1989catastrophic}. Classical mitigation strategies include regularization methods such as Elastic Weight Consolidation (EWC)~\cite{kirkpatrick2017overcoming}, Synaptic Intelligence~\cite{zenke2017continual}, and Learning without Forgetting~\cite{li2017learning}; replay-based methods such as Gradient Episodic Memory and exemplar replay~\cite{lopez2017gradient,rebuffi2017icarl}; and architectural methods such as Progressive Networks and parameter-isolation schemes~\cite{rusu2016progressive,mallya2018packnet}. These approaches provide strong baselines but typically require either explicit continual-learning objectives, task-specific storage, or architectural changes---leaving open the question of whether \emph{implicit} structural constraints, such as freezing most parameters, can achieve comparable forgetting reduction without dedicated anti-forgetting machinery.

\subsection{Parameter-Efficient Fine-Tuning}

PEFT methods adapt pretrained models while updating a small number of task-specific parameters. \LoRA~\cite{hu2021lora} is the most widely used low-rank approach. Subsequent variants include AdaLoRA~\cite{zhang2023adaptive}, DoRA~\cite{liu2024dora}, RS-LoRA~\cite{kong2024rslora}, and quantization-aware approaches such as QLoRA and LoftQ~\cite{dettmers2023qlora,li2024loftq}. These methods are primarily motivated by efficiency, but their frozen-backbone structure also makes them natural candidates for sequential adaptation. However, the PEFT literature overwhelmingly evaluates these methods on single-task accuracy; their behavior under \emph{sequential} adaptation---and specifically whether the frozen backbone itself drives forgetting reduction---remains systematically under-studied.

\subsection{PEFT for Continual Learning}

Several recent papers study \LoRA or related PEFT mechanisms directly in continual-learning settings. O-LoRA enforces orthogonality across low-rank updates \cite{wang2023olora}. Prompt-based methods such as L2P, DualPrompt, and CODA-Prompt adapt prompts instead of the full backbone \cite{wang2022l2p,wang2022dualprompt,smith2023codaprompt}. InfLoRA analyzes interference-free low-rank adaptation \cite{liang2024inflora}. More recent work includes C-LoRA \cite{zhang2025clora}, LoRI \cite{zhang2025lori}, replay combined with \LoRA for NLU \cite{borhanifard2025replay_lora}, drift-resistant \LoRA subtraction in exemplar-free continual learning \cite{liu2025lora_subtraction}, and a 2026 geometric account of forgetting in low-rank adaptation \cite{steele2026subspace_geometry}.

Our paper differs from the above work on three axes. First, we study \emph{standard} \LoRA without any continual-learning-specific modifications, isolating the implicit frozen-backbone effect. Second, we enforce strict protocol separation between confirmatory and exploratory evidence, preventing mixed-protocol overclaiming. Third, we connect the empirical result to a mechanistic interpretation via freezing ablations and representation probes, rather than only reporting aggregate forgetting scores.

\subsection{Task-Invariant Representations and Mechanistic Explanations}

Multi-task and transfer-learning research has long suggested that stable shared features can improve robustness across tasks, while task-specific drift in deeper layers can produce interference. Recent continual-learning work on PEFT increasingly frames the problem in terms of geometric separation, orthogonality, replay, or constrained subspaces rather than only aggregate forgetting scores. Our manuscript takes a narrower step in that direction. We do not claim a complete mechanistic proof, but we do test whether the repository's strongest evidence is more consistent with a frozen-backbone account than with a purely low-rank one. The language of task-invariant representations is therefore used here as an interpretation of the observed freezing pattern, not as a fully established theorem about all \LoRA behavior.

\section{Experimental Methodology}
\label{sec:methodology}

\subsection{Sequential Protocol and Forgetting Metric}

Our main task order is RTE $\rightarrow$ MRPC $\rightarrow$ CoLA $\rightarrow$ SST-2, using GLUE \cite{wang2018glue}. After each task is trained, we evaluate the current model on every task seen so far. For a task $T_i$, forgetting is the drop between its post-training peak and its accuracy after later tasks:
\begin{equation}
\Forget(T_i) = \text{Acc}_i(\text{after training } T_i) - \text{Acc}_i(\text{after training } T_j), \quad j > i.
\end{equation}
Average forgetting over prior tasks is
\begin{equation}
\overline{\Forget} = \frac{1}{N-1}\sum_{i=1}^{N-1} \Forget(T_i).
\end{equation}
Positive values indicate forgetting; negative values indicate backward transfer.

\subsection{Evidence Tiers and Protocol Taxonomy}

The central methodological choice in this paper is to separate confirmatory runs from exploratory ones. Earlier experiments in the repository used \texttt{validation[:200]} slices and were useful for screening, but they are not merged with the confirmatory statistics reported here. In particular, the older BERT seeds 42--44 are \emph{not} combined with the full-validation reruns 45--49.

\begin{table}[t]
\centering
\caption{Protocol taxonomy used in this paper. Only the confirmatory BERT row supplies the main headline result.}
\label{tab:protocol_taxonomy}
\footnotesize
\setlength{\tabcolsep}{3pt}
\begin{tabular}{@{}L{2.3cm}L{2.3cm}L{2.0cm}L{2.2cm}C{1.0cm}@{}}
\toprule
\textbf{Study} & \textbf{Role} & \textbf{Validation} & \textbf{Training caps} & \textbf{Seeds} \\
\midrule
\shortstack[l]{BERT full FT\\vs.\ \LoRA} & Confirmatory & Full validation & \shortstack[l]{SST-2 capped\\at 2,000} & 45--49 \\
\shortstack[l]{RoBERTa\\full FT\\vs.\ \LoRA} & Secondary & Full validation & \shortstack[l]{SST-2 capped\\at 2,000} & 42--44 \\
EWC baseline & Secondary & Full validation & \shortstack[l]{SST-2 capped\\at 2,000} & 42--44 \\
\shortstack[l]{6-task\\\LoRA\\extension} & Secondary & Full validation & \shortstack[l]{SST-2,\\QNLI capped\\at 2,000} & 42--44 \\
\shortstack[l]{Rank/module/\\quant.\ sweeps} & Exploratory & \shortstack[l]{\texttt{validation}\\\texttt{[:200]}} & Screening setup & 42--44 \\
\shortstack[l]{Fine-grained\\freezing\\ablation} & Exploratory & Full validation & \shortstack[l]{SST-2 capped\\at 2,000} & 42--44 \\
\bottomrule
\end{tabular}
\end{table}

\subsection{Models, Hyperparameters, and Statistics}

The confirmatory study uses BERT-base-uncased with either full fine-tuning or \LoRA ($r=8$, query/value modules). Secondary experiments use RoBERTa-base, EWC, 6-task sequential adaptation, and freezing ablations under the settings summarized in \cref{tab:protocol_taxonomy}. Across these runs, training uses three epochs, learning rate $2 \times 10^{-5}$, batch size 16, AdamW, warmup 100 steps, and weight decay 0.01. We report mean and standard deviation over seeds, and for the main BERT comparison we compute a paired $t$-test and Cohen's $d_z$ directly from the confirmatory reruns.

\subsection{Interpretive Scope}

The framing in this paper distinguishes between evidence that informs mechanism and a full mechanistic proof. The confirmatory BERT reruns establish that standard \LoRA strongly reduces average forgetting under the audited protocol. The freezing studies, classifier-only baselines, and shallow-adapter variants then ask which architectural constraint best matches that reduction. This lets us compare a frozen-backbone hypothesis against a weaker low-rank-only story. It also forces us to be explicit about what the current paper still does \emph{not} establish decisively: a uniquely causal mechanism, a gradient-geometry explanation that survives alternative probes, or long-horizon continual adaptation under matched seeds.

\subsection{Companion Mechanism Probes}

To test the frozen-backbone interpretation more directly, we additionally use a verified companion experiment repository that measures task similarity in representation space. These probes are kept analytically separate from the confirmatory BERT forgetting statistics: they use partly different architectures (GPT-2 and RoBERTa), sometimes fewer tasks, and were designed to discriminate between mechanism hypotheses rather than to extend the main five-seed benchmark. In these probes, \emph{task similarity} denotes cosine similarity between task centroids extracted from hidden-state activations, with higher values indicating that sequentially adapted tasks continue to occupy a more shared representation space. We use these studies only as explanatory evidence about freezing, gradual unfreezing, orthogonality, and layer-wise drift.

\begin{figure}[t]
\centering
\includegraphics[width=\linewidth]{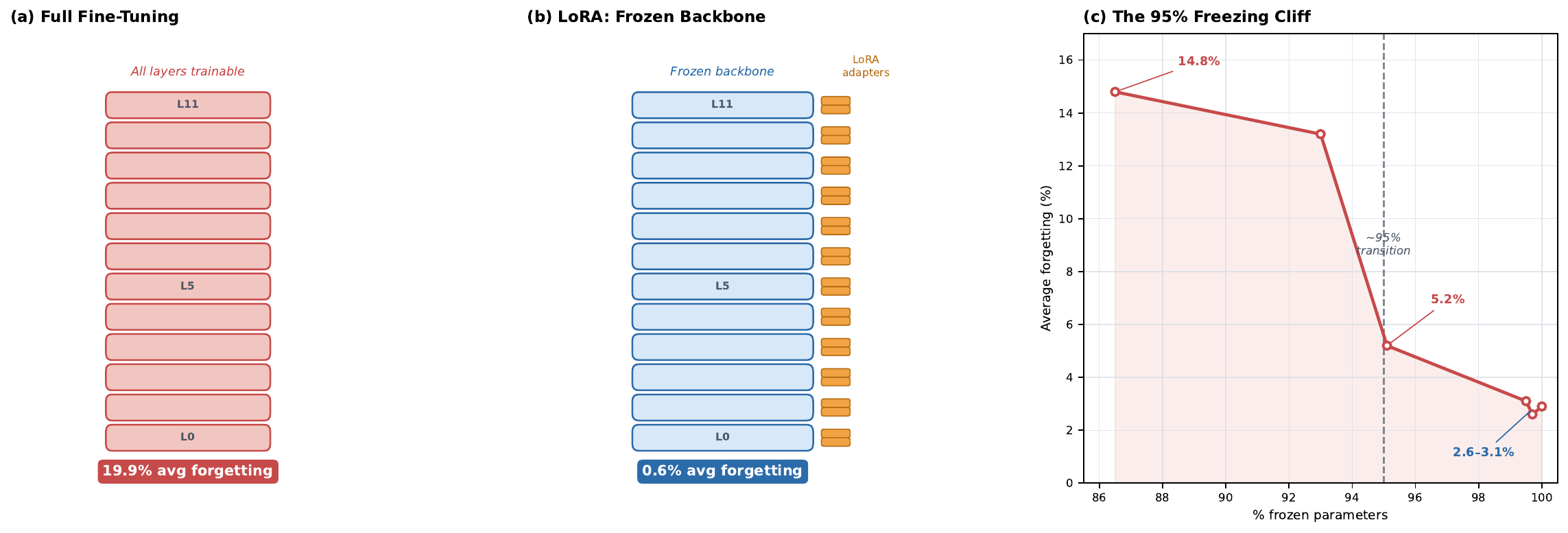}
\caption{Paper overview. Left: full fine-tuning leaves the entire backbone trainable and suffers high forgetting under unconstrained drift. Middle: \LoRA freezes the backbone and routes task-specific adaptation through small low-rank updates, preserving a more stable shared scaffold. Right: the empirical freezing ablation shows a sharp forgetting drop once the frozen fraction crosses roughly 95\%.}
\label{fig:paper_overview}
\end{figure}

\Cref{fig:paper_overview} summarizes the paper's central argument before the detailed results: the empirical gap between full fine-tuning and \LoRA is large, and the freezing transition suggests that the main driver is strong backbone preservation rather than unconstrained parameter drift.

\section{Main Results}
\label{sec:results}

\subsection{Confirmatory BERT-Base Result}

The cleanest evidence in this paper comes from the five full-validation BERT reruns. Under this consistent protocol, full fine-tuning produces substantial forgetting:
\begin{equation}
\boxed{\text{Full FT: } 19.9\% \pm 4.8\%}\,,
\end{equation}
while \LoRA yields much lower average forgetting:
\begin{equation}
\boxed{\LoRA\text{: } 0.6\% \pm 1.4\%}\,.
\end{equation}
This corresponds to a $97.2\%$ relative reduction. The paired comparison also favors \LoRA under this protocol ($t(4)=6.99$, $p = 0.002$, Cohen's $d_z = 3.12$).

\begin{figure}[t]
\centering
\includegraphics[width=0.62\linewidth]{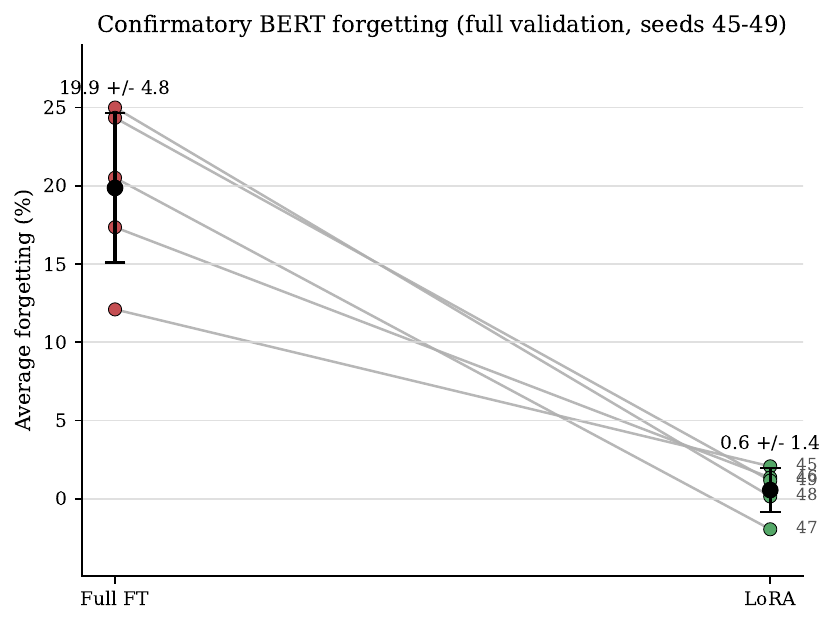}
\caption{Seed-paired confirmatory BERT forgetting comparison from the audited full-validation reruns (seeds 45--49).}
\label{fig:confirmatory_bert_main}
\end{figure}

\begin{table}[t]
\centering
\caption{Confirmatory BERT-base forgetting statistics from the full-validation reruns (seeds 45--49). Earlier subset-validation runs are intentionally excluded.}
\label{tab:main_forgetting}
\begin{tabular}{lccccc}
\toprule
\textbf{Method} & \textbf{Mean} & \textbf{Std} & \textbf{Min} & \textbf{Max} & \textbf{Reduction} \\
\midrule
Full fine-tuning & 19.9\% & 4.8\% & 12.1\% & 25.0\% & --- \\
\LoRA ($r\!=\!8$, qv) & \textbf{0.6\%} & \textbf{1.4\%} & $-2.0\%$ & 2.1\% & \textbf{97.2\%} \\
\bottomrule
\end{tabular}
\end{table}

\begin{figure}[t]
\centering
\begin{subfigure}[t]{0.48\linewidth}
  \centering
  \includegraphics[width=\linewidth]{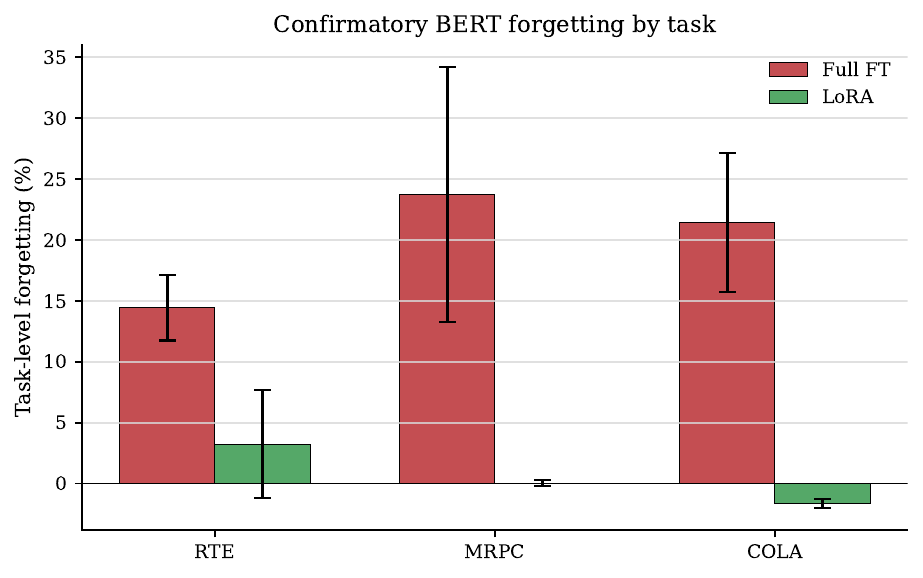}
  \caption{Task-level forgetting. \LoRA reduces forgetting on all previously learned tasks.}
  \label{fig:confirmatory_tasklevel}
\end{subfigure}\hfill
\begin{subfigure}[t]{0.48\linewidth}
  \centering
  \includegraphics[width=\linewidth]{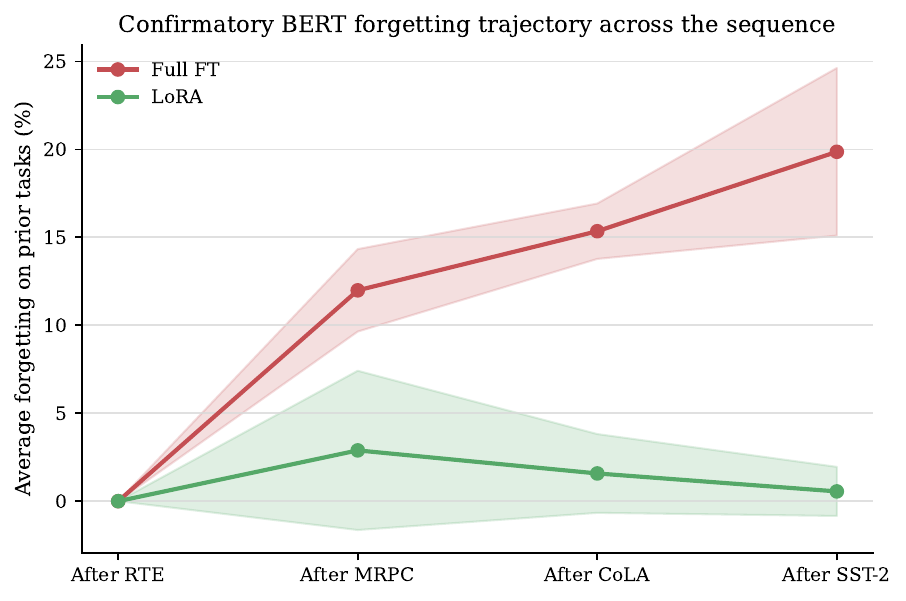}
  \caption{Stage-wise average forgetting trajectory across the confirmatory BERT sequence.}
  \label{fig:confirmatory_trajectory}
\end{subfigure}
\caption{Task-level and stage-wise forgetting in the confirmatory BERT reruns (seeds 45--49). (a) \LoRA reduces forgetting on every individual task. (b) The gap between methods widens progressively as more tasks are added.}
\label{fig:confirmatory_tasklevel_trajectory}
\end{figure}

Task-level averages show the same qualitative pattern. Under full fine-tuning, average forgetting is largest on MRPC (23.7 percentage points), followed by CoLA (21.4) and RTE (14.4). Under \LoRA, MRPC forgetting is approximately zero on average, RTE forgetting is 3.3 points, and CoLA shows slight backward transfer. \Cref{fig:confirmatory_tasklevel} shows that the reduction is not confined to the aggregate statistic: forgetting is lower on each previously learned task, with the largest absolute gaps on MRPC and CoLA. The confirmatory reruns also show an accuracy trade-off: final average accuracy is 64.5\% for full fine-tuning versus 59.5\% for \LoRA.

\begin{figure}[t]
\centering
\includegraphics[width=0.65\linewidth]{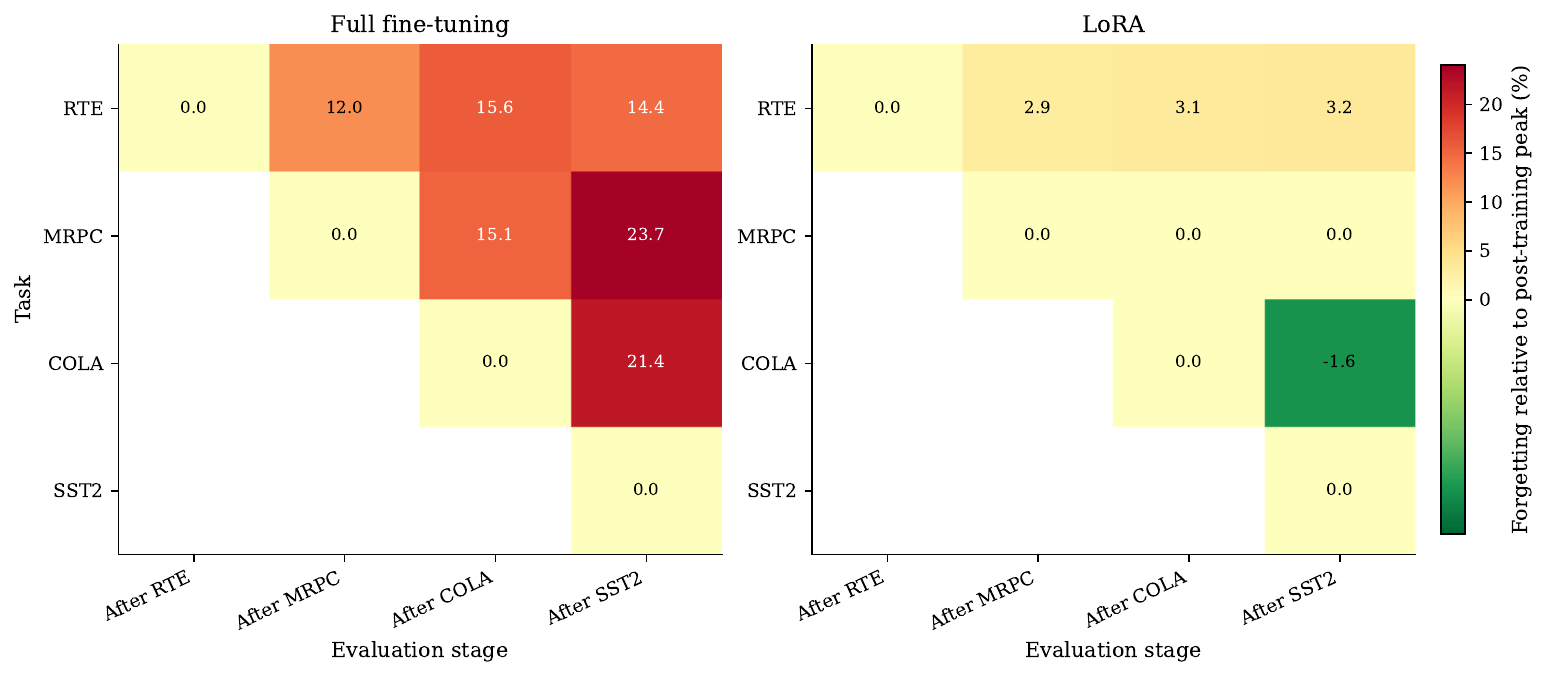}
\caption{Mean task-by-stage forgetting in the confirmatory BERT reruns. Each cell reports forgetting relative to that task's post-training peak, averaged across seeds 45--49.}
\label{fig:confirmatory_heatmap}
\end{figure}

\Cref{fig:confirmatory_heatmap} makes the task-by-stage pattern explicit. Under full fine-tuning, forgetting accumulates quickly for MRPC and CoLA after later tasks. The \LoRA heatmap remains near zero except for modest RTE drift and slight CoLA backward transfer.

\subsection{Exploratory Configuration Context}

Earlier screening sweeps over rank, module targeting, and quantization used \texttt{validation[:200]} and are therefore treated as exploratory only. Those runs suggested that $r=8$ was a reasonable parameter-efficiency starting point, that targeting more linear layers could improve task accuracy, and that 4-bit quantization preserved accuracy in the screening setup while reducing memory footprint. Across the appendix sweeps, the configuration story is internally consistent: larger rank improves screening accuracy at rising parameter cost (\cref{fig:appendix_rank}), all-linear targeting improves screening accuracy relative to query/value-only targeting (\cref{fig:appendix_modules}), and 4-bit quantization preserves the same qualitative ordering while lowering memory (\cref{fig:appendix_quantization}). We retain these sweeps as motivation for future confirmatory work, not as headline forgetting evidence. The detailed exploratory plots are included in \cref{sec:appendix_exploratory_figures}.

\section{Secondary Evidence}
\label{sec:secondary}

\subsection{RoBERTa Replication}

The full-validation RoBERTa replication supports the same qualitative picture: forgetting remains much lower with standard \LoRA than with full fine-tuning.

\begin{table}[t]
\centering
\caption{Secondary full-validation RoBERTa-base replication ($n=3$ seeds).}
\label{tab:roberta}
\small
\setlength{\tabcolsep}{4pt}
\begin{tabular}{@{}L{2.9cm}C{2.3cm}C{2.5cm}C{2.0cm}@{}}
\toprule
\textbf{Method} & \textbf{Forgetting} & \textbf{Final Avg. Accuracy} & \textbf{Reduction} \\
\midrule
Full fine-tuning & 24.2\% $\pm$ 2.2\% & 64.8\% & --- \\
\LoRA ($r\!=\!8$, qv) & \textbf{1.7\% $\pm$ 1.0\%} & 59.3\% & \textbf{92.8\%} \\
\bottomrule
\end{tabular}
\end{table}

\subsection{EWC Baseline}

We also ran a separate full-validation EWC baseline on the same 4-task order. The strongest EWC setting in the repository reaches $15.5\% \pm 1.4\%$ forgetting with 67.0\% final average accuracy. This is higher forgetting than the confirmatory \LoRA reruns, but we treat the comparison as contextual rather than paired, because the EWC runs use a different seed set.

\begin{table}[t]
\centering
\caption{Secondary EWC baseline under the same task order and training setup. The BERT \LoRA row is included only as confirmatory context, not as a paired same-seed comparison.}
\label{tab:cl_comparison}
\small
\setlength{\tabcolsep}{4pt}
\begin{tabular}{@{}L{3.1cm}C{0.9cm}C{2.2cm}C{2.5cm}@{}}
\toprule
\textbf{Method} & \textbf{$n$} & \textbf{Forgetting} & \textbf{Final Avg. Accuracy} \\
\midrule
BERT \LoRA (confirmatory) & 5 & \textbf{0.6\% $\pm$ 1.4\%} & 59.5\% \\
EWC ($\lambda=100$) & 6 & 16.4\% $\pm$ 1.2\% & 66.2\% \\
EWC ($\lambda=1000$) & 3 & 16.2\% $\pm$ 1.2\% & 66.2\% \\
EWC ($\lambda=5000$) & 3 & 15.5\% $\pm$ 1.4\% & 67.0\% \\
\bottomrule
\end{tabular}
\end{table}

\begin{figure}[t]
\centering
\includegraphics[width=0.68\linewidth]{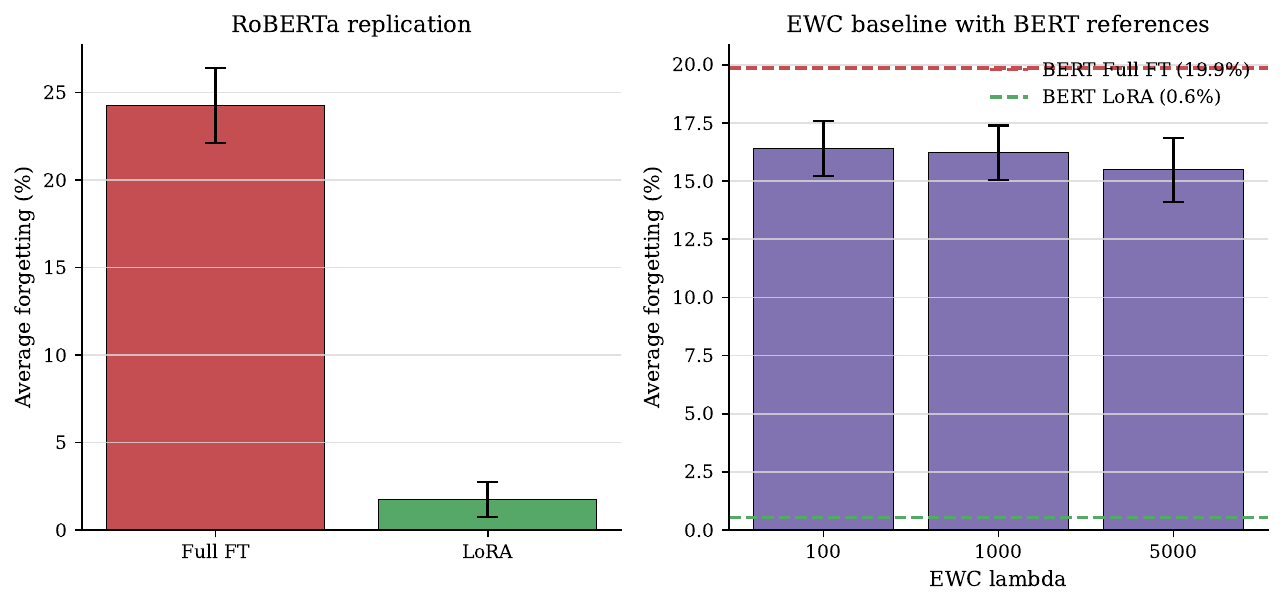}
\caption{Secondary cross-checks. Left: RoBERTa full fine-tuning versus \LoRA. Right: EWC forgetting by $\lambda$ with confirmatory BERT reference lines.}
\label{fig:secondary_cross_checks}
\end{figure}

\subsection{Six-Task Extension}
\label{sec:6task}

The 6-task extension adds QNLI and WNLI. Average forgetting remains low relative to full fine-tuning, but the task-level breakdown is heterogeneous enough that we do not treat this section as clean scalability evidence.

\begin{table}[t]
\centering
\caption{LoRA under 4-task and 6-task full-validation protocols. The 6-task result has low average forgetting but large MRPC forgetting.}
\label{tab:6task}
\small
\begin{tabular}{@{}L{3.3cm}C{0.8cm}C{2.2cm}C{2.3cm}C{2.0cm}@{}}
\toprule
\textbf{Setting} & \textbf{$n$} & \textbf{Avg. Forgetting} & \textbf{Final Avg. Accuracy} & \textbf{MRPC Forgetting} \\
\midrule
\shortstack[l]{4-task BERT\\\LoRA reference} & 5 & 0.6\% $\pm$ 1.4\% & 59.5\% & $-0.3$ to $0.5$ pts \\
6-task BERT \LoRA & 3 & 5.6\% $\pm$ 0.8\% & 53.8\% & 35.8--37.5 pts \\
\bottomrule
\end{tabular}
\end{table}

\begin{figure}[t]
\centering
\begin{subfigure}[t]{0.48\linewidth}
  \centering
  \includegraphics[width=\linewidth]{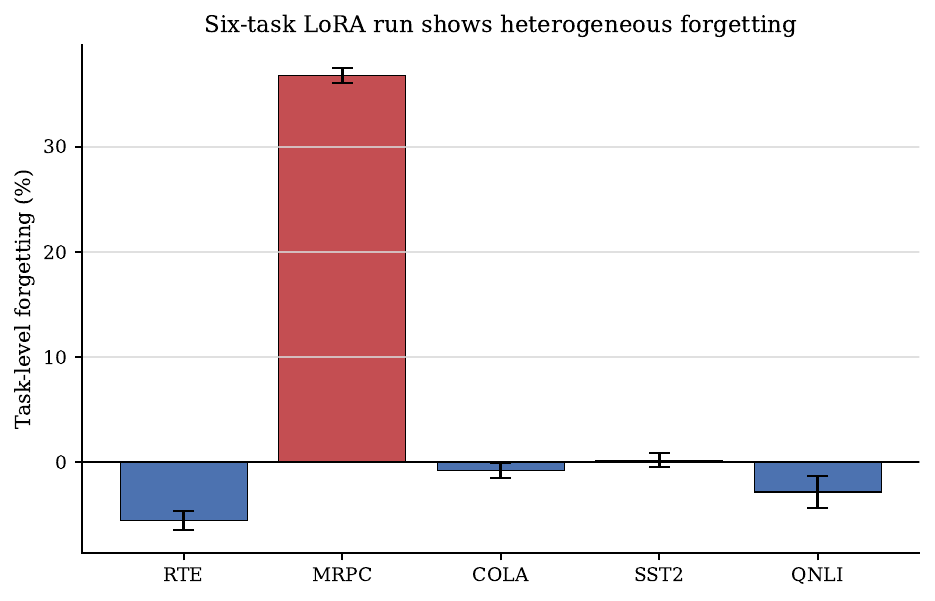}
  \caption{Task-level forgetting. MRPC is the clear outlier.}
  \label{fig:six_task_heterogeneity}
\end{subfigure}\hfill
\begin{subfigure}[t]{0.48\linewidth}
  \centering
  \includegraphics[width=\linewidth]{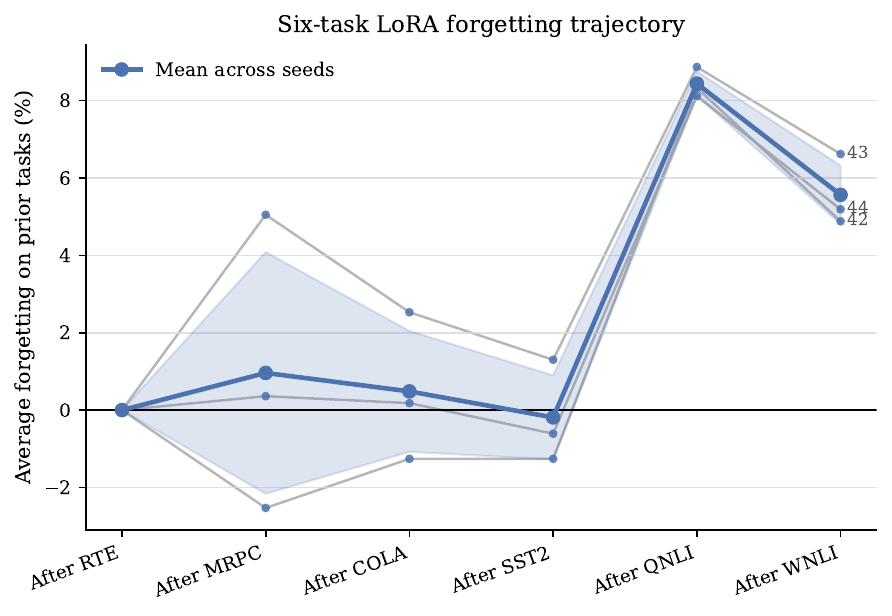}
  \caption{Average-forgetting trajectory. Faint lines show individual seeds.}
  \label{fig:six_task_trajectory}
\end{subfigure}
\caption{Six-task \LoRA extension. (a) MRPC dominates task-level forgetting despite low average values. (b) Forgetting stays near zero through SST-2, then jumps sharply once QNLI is introduced.}
\label{fig:six_task_combined}
\end{figure}

A stage-wise view clarifies where the average changes occur. Forgetting stays close to zero through SST-2, then jumps sharply once QNLI is introduced and remains elevated after WNLI (\cref{fig:six_task_trajectory}).

\section{Freezing and Representation Evidence for a Frozen-Backbone Hypothesis}
\label{sec:ablation_freezing}

The fine-grained freezing ablation uses full validation and therefore carries more weight than the screening sweeps, but we still treat it as secondary evidence rather than as a second confirmatory headline. Its purpose is not to replace the main BERT comparison; instead, it asks which part of the \LoRA design best matches the forgetting reduction. The main pattern is clear: forgetting decreases markedly once more than 95\% of parameters are frozen.

\begin{table}[t]
\centering
\caption{Exploratory full-validation freezing ablation ($n=3$ seeds). The pattern suggests a marked drop in forgetting once freezing exceeds roughly 95\%, but we do not interpret this as identifying a unique causal mechanism.}
\label{tab:freezing_ablation}
\footnotesize
\setlength{\tabcolsep}{3pt}
\begin{tabular}{@{}L{3.1cm}C{1.2cm}C{1.5cm}C{1.8cm}L{2.4cm}@{}}
\toprule
\textbf{Configuration} & \textbf{Frozen \%} & \textbf{Avg. Forgetting} & \shortstack[c]{\textbf{Trainable}\\ \textbf{Params}} & \textbf{Role} \\
\midrule
\shortstack[l]{Freeze layers 0--9,\\train 10--11} & 86.5\% & 14.8\% & 14.8M & \shortstack[l]{High-forgetting\\reference} \\
\shortstack[l]{Freeze layers 0--10,\\train 11} & 93.0\% & 13.2\% & 7.7M & \shortstack[l]{High-forgetting\\reference} \\
\shortstack[l]{Freeze 0--10 +\\layer-11 attention} & 95.1\% & 5.2\% & 5.3M & Marked drop \\
Freeze all 12 layers & 99.5\% & 3.1\% & 592K & \shortstack[l]{Classifier-only\\baseline} \\
\LoRA on layer 11 only & 100.0\% & 2.9\% & 26K & \shortstack[l]{Adapter\\baseline} \\
\LoRA on layers 10--11 & 100.0\% & 2.9\% & 51K & \shortstack[l]{Adapter\\baseline} \\
\LoRA on all layers & 99.7\% & \textbf{2.6\%} & 296K & Standard \LoRA \\
\bottomrule
\end{tabular}
\end{table}

\begin{figure}[t]
\centering
\includegraphics[width=0.62\linewidth]{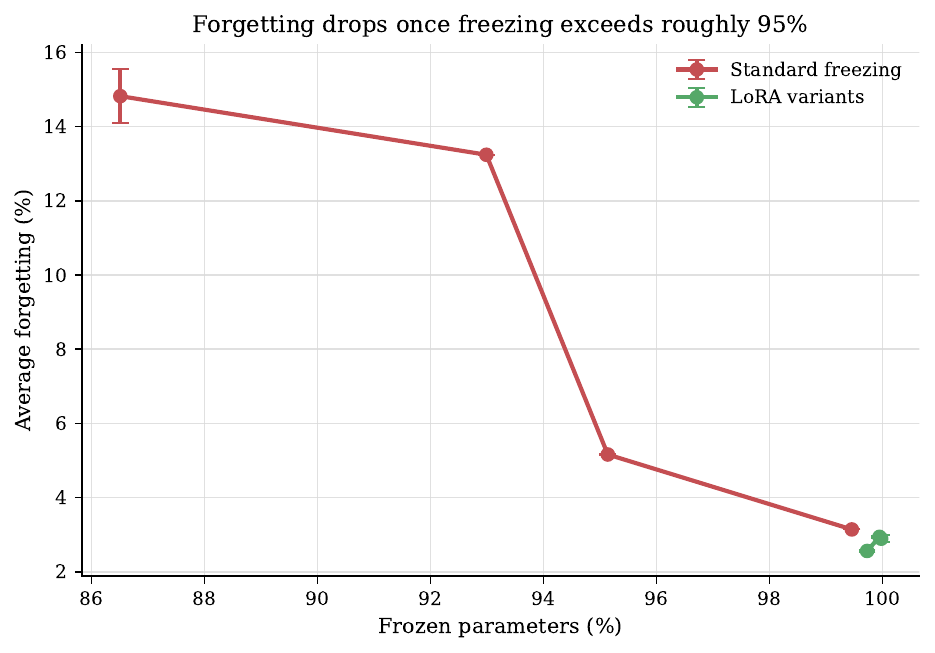}
\caption{Fine-grained freezing ablation. Average forgetting falls sharply once freezing exceeds roughly 95\% of parameters.}
\label{fig:freezing_transition}
\end{figure}

This ablation is consistent with a frozen-backbone interpretation of the main result, but it does not by itself prove that attention weights are the unique locus of interference. We therefore frame it as suggestive evidence about where forgetting becomes easier to control, not as a mechanistic proof.

Read together, the freezing rows suggest a more specific interpretation than the original empirical paper alone could support. High-forgetting partially unfrozen configurations (86.5\% and 93.0\% frozen) behave much more like full fine-tuning than like standard \LoRA, whereas classifier-only and shallow-adapter baselines remain close to the standard \LoRA regime once the frozen fraction approaches 100\%. This pattern is more naturally explained by a strong frozen-backbone constraint than by low-rank parameterization alone. In that sense, the repository's evidence points toward a task-invariant-subspace hypothesis: when almost all backbone parameters are fixed, later tasks appear to reuse a more stable shared feature scaffold instead of overwriting it.

The companion mechanism probes strengthen this interpretation by measuring task similarity directly rather than inferring it only from forgetting curves. In a four-task GPT-2 probe, a fully frozen backbone reaches task similarity $0.9988$ versus $0.8935$ for full fine-tuning while retaining competitive mean accuracy ($0.671$ vs.\ $0.701$). A gradual-unfreezing probe then shows monotonic erosion as more layers are released: task similarity falls from $0.9982$ in the head-only regime to $0.9943$ and $0.9911$ under progressively broader schedules, then collapses to $0.8542$ under full fine-tuning. A dedicated layer-wise representation analysis shows that both regimes remain highly similar through most layers, but full fine-tuning exhibits its clearest drop at the final layer ($0.9902$ versus $0.9973$ for \LoRA). Orthogonal-regularization sweeps remain effectively flat ($0.9775$--$0.9778$), suggesting that generic update decorrelation alone does not recreate the same preservation effect. The broader direction is also visible in RoBERTa ($0.9321$ vs.\ $0.6383$), GPT-2 Medium ($0.9994$ vs.\ $0.9967$), and an eight-seed GPT-2 validation where \LoRA maintains higher task similarity than full fine-tuning ($0.9973 \pm 0.0016$ vs.\ $0.9897 \pm 0.0069$, $p = 0.013$; see \cref{fig:appendix_companion_robustness} for the compact robustness summary).

\begin{figure}[t]
\centering
\includegraphics[width=0.78\linewidth]{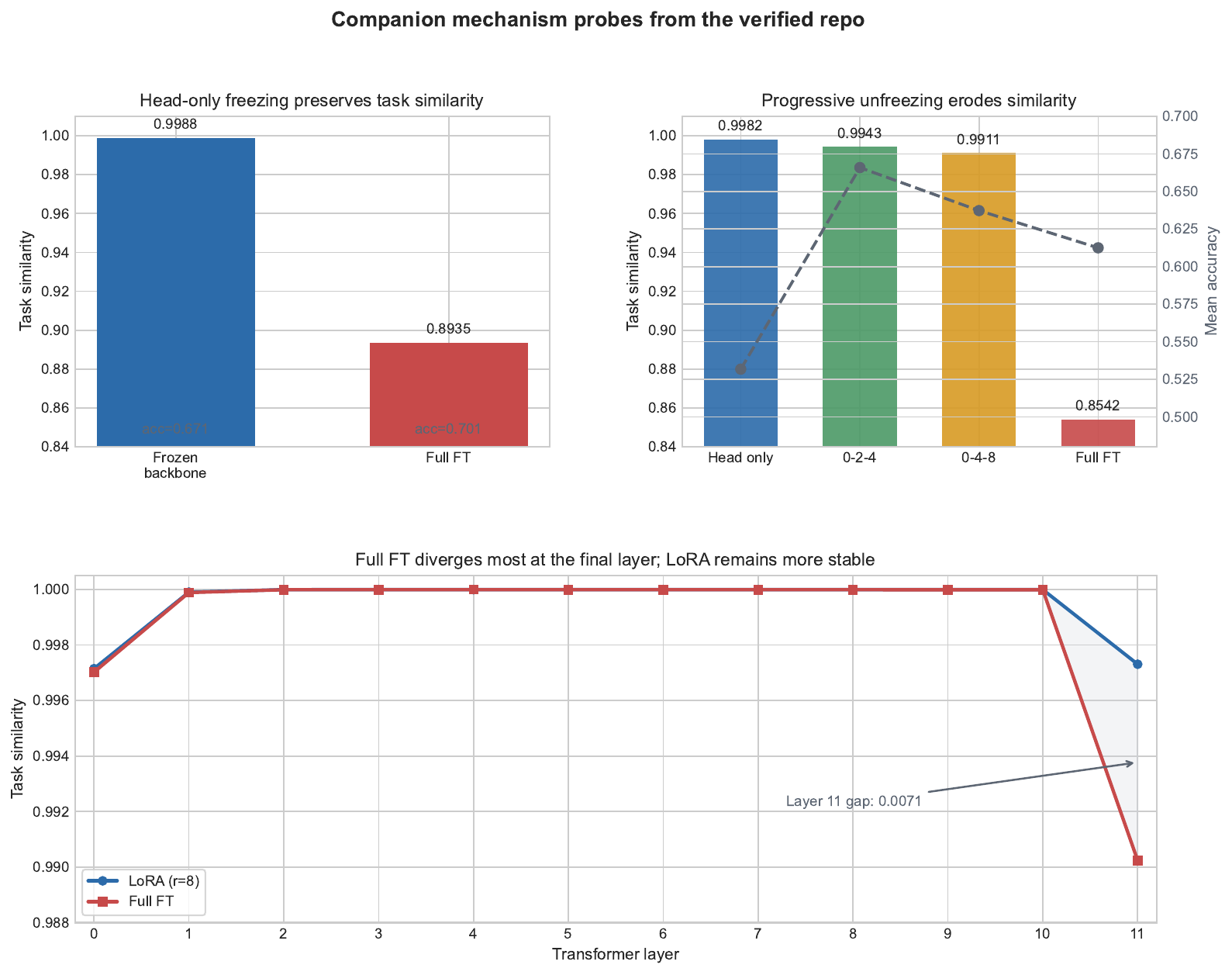}
\caption{Companion mechanism probes from the verified second repository. Top left: direct task-similarity comparison between a frozen backbone and full fine-tuning in a four-task GPT-2 probe. Top right: task similarity degrades steadily as more layers are unfrozen. Bottom: layer-wise task similarity remains high in both regimes, but full fine-tuning shows its largest drop at the final transformer layer. These probes are exploratory and architecture-mismatched to the main BERT benchmark, so they strengthen interpretation rather than change the confirmatory headline result.}
\label{fig:companion_mechanism_probes}
\end{figure}

\section{Computational Footprint}

The repository's directly audited profiling logs are limited to one exploratory 4-bit \LoRA configuration, so we report them only as deployment context and not as a cross-method speedup claim. In that setup, training takes $11.95 \pm 1.12$ seconds per epoch, peak memory is $0.46 \pm 0.11$ GB, inference latency is $53.72 \pm 4.62$ ms, throughput is 18.74 samples/s, and the trainable parameter count is 296,450 (0.44\% of the quantized model). These values should be read together with the exploratory quantization figure in the appendix: they show that the all-linear/4-bit configuration can operate at sub-gigabyte memory scale, but they are peripheral to the paper's main empirical and mechanistic claims.

\section{Discussion}
\label{sec:discussion}

\subsection{What the Confirmatory Result Does and Does Not Show}

The strongest claim supported by the repository is that standard \LoRA substantially reduces \emph{average} forgetting in the audited 4-task BERT setting. That claim is robust to the stricter protocol audit: even after discarding the older subset-validation seeds, the confirmatory reruns still show a large reduction in forgetting. However, the result does not imply that forgetting is solved in general, nor that every task is uniformly protected.

\subsection{Accuracy-Forgetting Trade-off}

The confirmatory reruns show that forgetting reduction is not free. Full fine-tuning reaches higher final average accuracy than \LoRA (64.5\% vs.\ 59.5\%), even after forgetting. The 6-task extension reinforces the same point: average forgetting can remain low while individual tasks deteriorate sharply. For practical continual-learning deployments, average forgetting should therefore be reported together with task-level breakdowns and final average accuracy.

\subsection{Limitations}

We identify five principal limitations, each with its scope of impact and a path toward resolution:

\begin{itemize}[leftmargin=*,itemsep=4pt]
\item \textbf{Task and architecture scope.} The confirmatory claim rests on BERT-base encoder-only classification over four GLUE tasks. This restricts direct generalization to decoder models, generative tasks, or longer task sequences. \emph{Path forward:} extending the confirmatory protocol to GPT-2-scale decoders and $\geq$10-task sequences with matched seeds.

\item \textbf{Single \LoRA configuration.} The headline uses $r\!=\!8$ on query/value modules only. Different ranks, target modules, or variants (AdaLoRA, DoRA) may shift the forgetting--accuracy trade-off. \emph{Path forward:} a factorial sweep over rank $\times$ target module $\times$ \LoRA variant under the same full-validation protocol.

\item \textbf{Architecture mismatch in mechanism probes.} The companion task-similarity and layer-wise probes use GPT-2 and RoBERTa---architecturally distinct from the confirmatory BERT benchmark. They therefore strengthen interpretation rather than extend the confirmatory result. \emph{Path forward:} replicating the mechanism probes within the same BERT-base encoder used for the headline comparison.

\item \textbf{No replay or advanced CL baselines.} We compare only against EWC and vanilla full fine-tuning. Replay-augmented or recent PEFT-CL methods (O-LoRA, InfLoRA) are not included under a matched seed protocol. \emph{Path forward:} integrating these baselines into the same confirmatory framework.

\item \textbf{Screening data leakage risk.} The repository contains earlier \texttt{validation[:200]} sweeps. Although we do not pool these with confirmatory statistics, their existence requires transparent disclosure to avoid accidental mixed-protocol claims. \emph{Path forward:} archiving screening runs in a separate repository branch with explicit metadata tags.
\end{itemize}

\subsection{Frozen-Backbone Bottleneck Hypothesis}

The evidence in this paper supports a simple three-part design hypothesis rather than a completed mechanistic theory. First, strong freezing acts as an \emph{architectural bottleneck}: once most parameters are frozen, later tasks lose the degrees of freedom needed to freely rewrite earlier solutions. Second, that bottleneck appears to preserve a \emph{shared feature scaffold}. The freezing ablations suggest that forgetting falls most sharply when adaptation is pushed toward classifier-only or shallow-adapter regimes, which is exactly where most of the pretrained backbone is reused rather than rewritten. Third, the remaining plasticity becomes \emph{localized}. The system can still adapt, but the cost-benefit trade-off shifts toward limited task-specific heads or adapter paths rather than full-model drift.

This hypothesis is narrower than the mechanism paper's original claims, but it now rests on more than forgetting summaries alone. The companion repository contributes direct task-similarity, gradual-unfreezing, and layer-wise analyses that all point in the same direction: stronger freezing preserves a more shared representation space, while full fine-tuning induces substantially larger drift and concentrates its clearest divergence in the deepest layer. Even so, we still stop short of claiming a unique causal mechanism because the direct probes remain exploratory, use partly different architectures and task subsets, and do not isolate every alternative explanation.

\subsection{Interpretation}

The observed pattern is consistent with recent theoretical work that links forgetting in low-rank adaptation to the geometry of task subspaces \cite{steele2026subspace_geometry}. Our empirical findings also align qualitatively with the broader literature showing that task isolation, replay, orthogonality, or stronger freezing can all reduce interference \cite{wang2023olora,zhang2025clora,zhang2025lori,borhanifard2025replay_lora,liu2025lora_subtraction}. The companion probes sharpen that interpretation in two useful ways. First, orthogonal-regularization sweeps leave task similarity almost unchanged, which weakens a generic update-decorrelation explanation. Second, direct representation probes show that stronger freezing and smaller trainable subspaces move the system toward higher task similarity, while full fine-tuning produces the clearest late-layer divergence. The interpretation developed here is therefore stronger than a pure benchmark report but weaker than a full mechanistic proof: standard \LoRA appears to reduce forgetting largely because frozen-backbone constraints preserve a more task-invariant feature scaffold than full fine-tuning. We do not claim to have established the unique causal mechanism, only to have identified the frozen-trainable-subspace boundary as the best-supported explanation in the current repositories.

\section{Conclusion}
\label{sec:conclusion}

After separating confirmatory full-validation evidence from earlier exploratory runs, the central empirical result remains: in five BERT-base reruns, standard \LoRA reduces average forgetting from $19.9\%$ to $0.6\%$. Secondary full-validation experiments on RoBERTa and EWC support the same qualitative story, while the 6-task, freezing, and companion representation probes show that the picture is more nuanced than a single average can capture.

The main lesson of this paper is therefore twofold. Standard \LoRA is a useful empirical baseline for sequential transformer-encoder adaptation, and the strongest current explanation for that robustness is the frozen-backbone constraint it imposes on later task learning. Protocol transparency still matters because mixing exploratory and confirmatory evidence can easily overstate certainty, but an equally important lesson is mechanistic: once direct task-similarity and layer-wise probes are added, the evidence increasingly points to representation preservation rather than low rank alone. We hope this paper is useful both as a controlled empirical result about low-rank adaptation and as a more interpretable account of why selective parameter freezing can matter in continual fine-tuning.

\section*{Acknowledgments}

This research received no external funding. Large language models were used to assist with language editing and manuscript drafting; all ideas, analyses, verification of claims, and final responsibility for the content remain with the author. The code, experiment configurations, and summary results supporting this study are available from the corresponding author upon reasonable request.

\bibliographystyle{plainnat}
\bibliography{references}

\appendix
\section{Exploratory Screening Figures}
\label{sec:appendix_exploratory_figures}

This appendix collects exploratory figures that informed configuration selection together with one companion robustness figure from the verified second repository. The first four figures use \texttt{validation[:200]} screening; training logs are limited to one exploratory 4-bit \LoRA configuration, so we report them only as deployment context and not as a cross-method speed comparison.

\begin{figure}[ht!]
\centering
\begin{subfigure}[t]{0.48\linewidth}
  \centering
  \includegraphics[width=\linewidth]{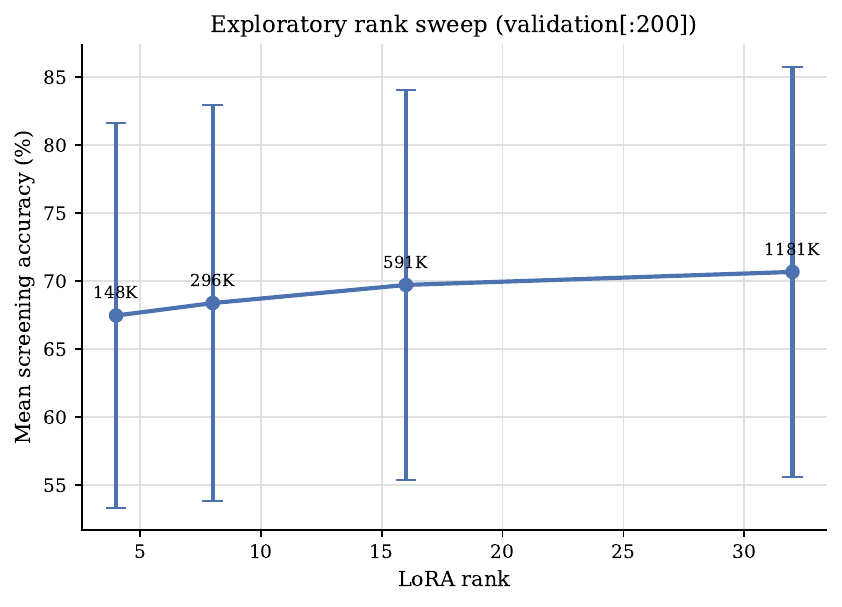}
  \caption{Rank sweep. Accuracy improves with larger rank but parameter count rises quickly.}
  \label{fig:appendix_rank}
\end{subfigure}\hfill
\begin{subfigure}[t]{0.48\linewidth}
  \centering
  \includegraphics[width=\linewidth]{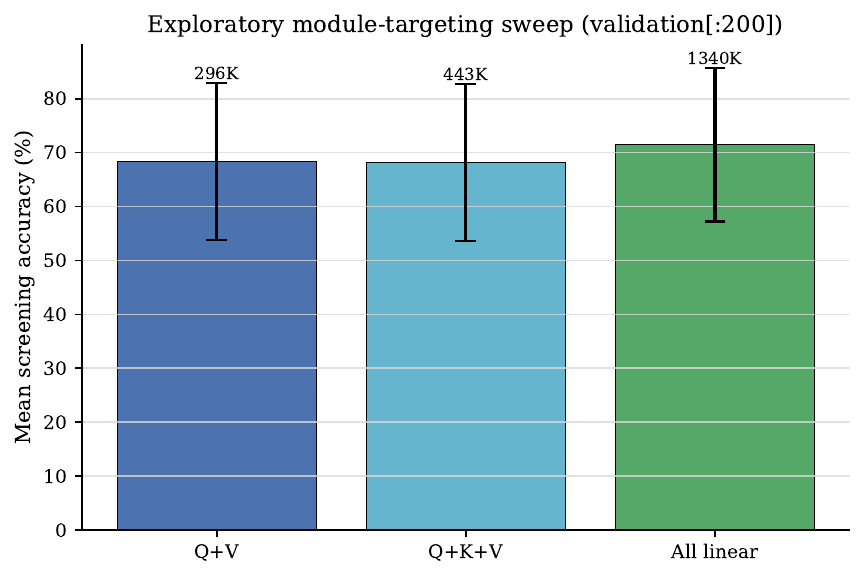}
  \caption{Module-targeting sweep. All-linear targeting improves accuracy over query/value.}
  \label{fig:appendix_modules}
\end{subfigure}
\caption{Exploratory configuration sweeps on \texttt{validation[:200]}.}
\label{fig:appendix_rank_modules}
\end{figure}

\begin{figure}[ht!]
\centering
\begin{subfigure}[t]{0.48\linewidth}
  \centering
  \includegraphics[width=\linewidth]{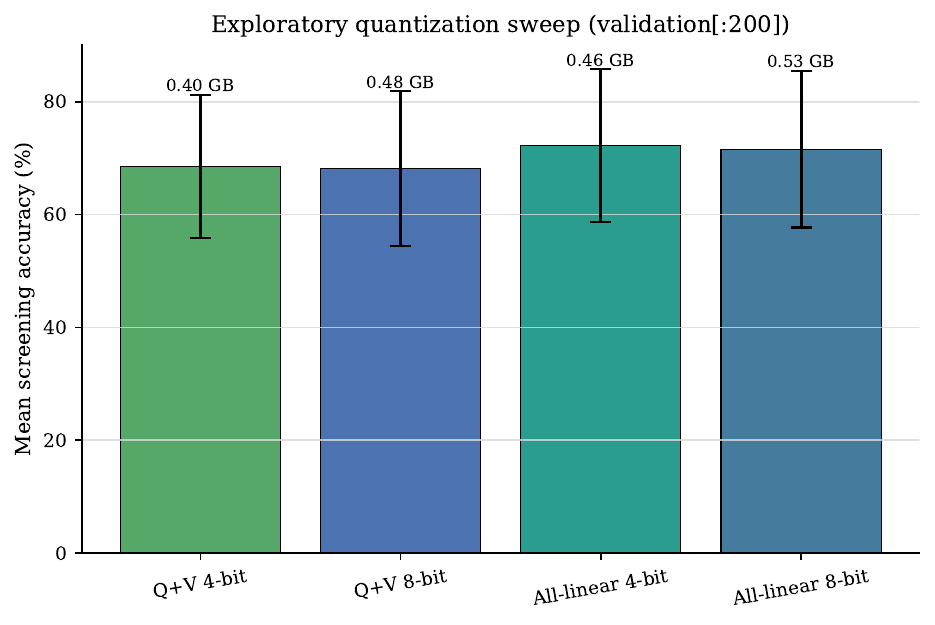}
  \caption{Quantization sweep. 4-bit preserves qualitative ranking while reducing memory.}
  \label{fig:appendix_quantization}
\end{subfigure}\hfill
\begin{subfigure}[t]{0.48\linewidth}
  \centering
  \includegraphics[width=\linewidth]{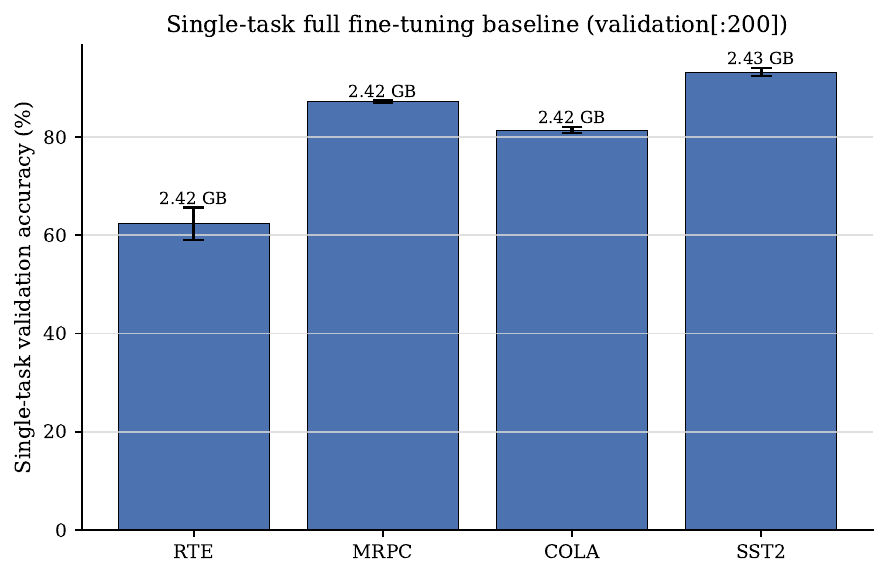}
  \caption{Single-task full fine-tuning baseline from the screening setup.}
  \label{fig:appendix_phase0}
\end{subfigure}
\caption{Exploratory quantization and baseline context from \texttt{validation[:200]}.}
\label{fig:appendix_quant_baseline}
\end{figure}

\begin{figure}[ht!]
\centering
\includegraphics[width=0.82\linewidth]{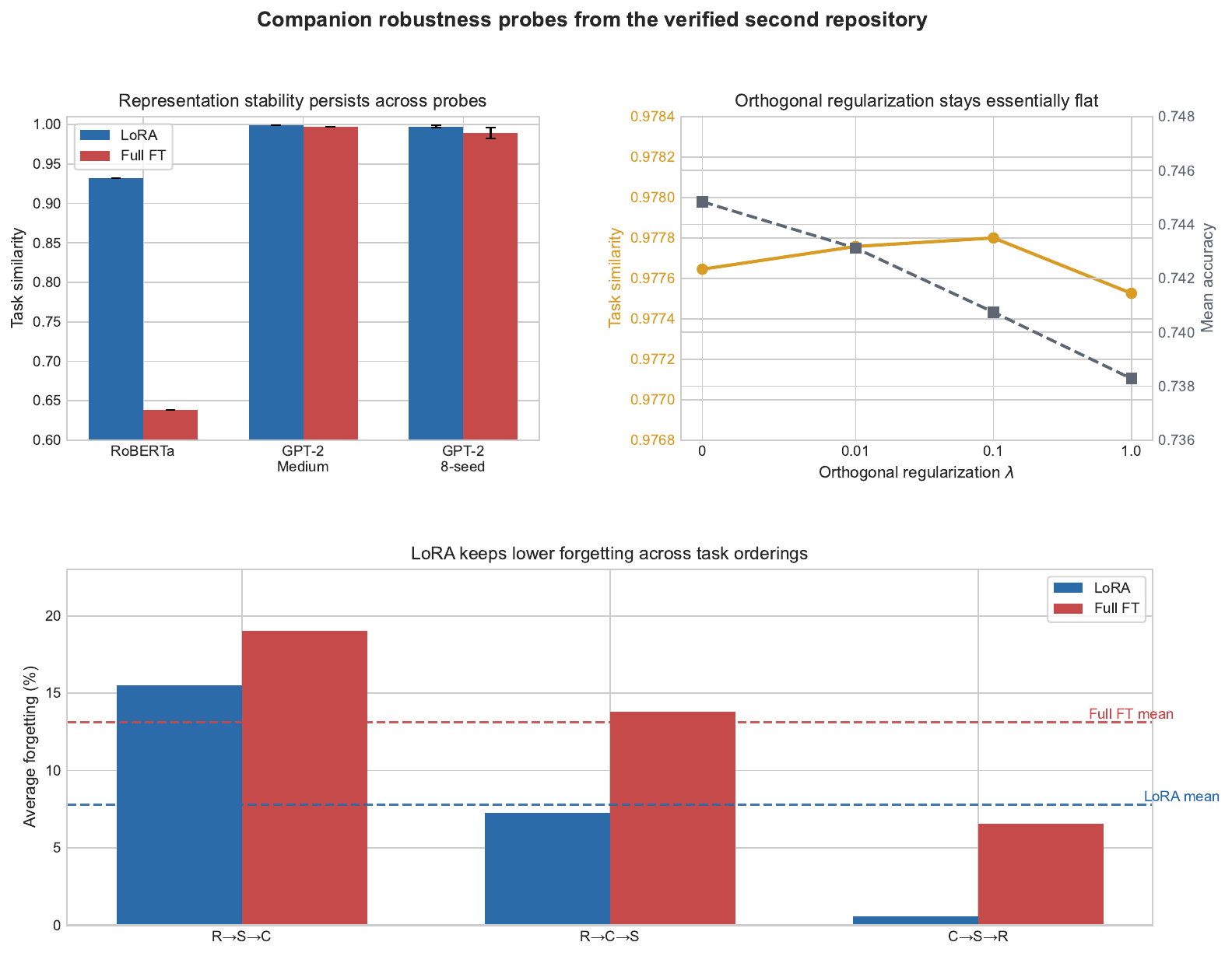}
\caption{Companion robustness probes. Left: task similarity across RoBERTa, GPT-2 Medium, and eight-seed GPT-2. Right: orthogonal regularization effect. Bottom: forgetting across task orderings.}
\label{fig:appendix_companion_robustness}
\end{figure}

\end{document}